\documentclass[10pt,letterpaper]{article}

\usepackage{cogsci}

\cogscifinalcopy 

\usepackage{graphicx}
\usepackage{pslatex}
\usepackage{apacite}
\usepackage{float} 
\title{Enhancing Creativity in Large Language Models through Associative Thinking Strategies}
 
\author{{\large \bf Pronita Mehrotra (pronita@mindantix.com)} \\
  MindAntix, WA \\
  \AND {\large \bf Aishni Parab (aishni@g.ucla.edu)} \\
  Department of Statistics, University of California Los Angeles \\
  \AND {\large \bf Sumit Gulwani (sumitg@microsoft.com)} \\
  Microsoft, Redmond, WA \\
  }

\begin{document}

\maketitle

\begin{abstract}
This paper explores the enhancement of creativity in Large Language Models (LLMs) like vGPT-4 through \textit{associative thinking}, a cognitive process where creative ideas emerge from linking seemingly unrelated concepts. Associative thinking strategies have been found to effectively help humans boost creativity. However, whether the same strategies can help LLMs become more creative remains under-explored. In this work, we investigate whether prompting LLMs to connect disparate concepts can augment their creative outputs. Focusing on three domains -- Product Design, Storytelling, and Marketing -- we introduce creativity tasks designed to assess vGPT-4's ability to generate original and useful content. By challenging the models to form novel associations, we evaluate the potential of associative thinking to enhance the creative capabilities of LLMs. Our findings show that leveraging associative thinking techniques can significantly improve the originality of vGPT-4's responses.

\end{abstract}

\section{Introduction}
Creativity, a cornerstone of human intelligence, plays a crucial role in our ability to adapt and innovate. Unlike conventional computational skills, creativity involves generating novel solutions and perspectives that are useful in solving complex, real-world problems. The current, widely accepted, definition of creativity includes two components - novelty and usefulness (see \cite{Runco2012a} for a historical account of this definition). For instance, Google’s PageRank algorithm was a creative solution since they used a \textit{unique} approach to rank sites based on importance and not just content matching, and it significantly \textit{improved} the quality of search results. Their inventive leap was not merely a result of linear problem-solving but emerged from combining the concept of academic citations (heavily cited articles are more authoritative sources) with websites -- two unrelated domains -- to create an automated way to rank websites \cite{Dyer2019a}.

Where do creative ideas come from? One perspective posits that humans store information in the brain in an associative nature where two concepts are connected by a particular link \cite{Kahneman2011a}. Creative insights often emerge by navigating unconventional paths within this associative network \cite{Benedek2012a, Michalko2000a}. For example, the concept of an ``apple'' may be connected to the concepts ``fruit'' (\textit{thing to category link}), ``red'' (\textit{thing to property link}), and ``snow white'' (\textit{thing to usage link; a poisoned apple is used in the popular fairy tale, Snow White}). Each of these different types of links exist under different strengths. For example, the link between ``apple'' and ``fruit'' may be the strongest among the three as it is a widely accepted concept, whereas ``apple-snow white'' may be connected by a weaker link for someone who is culturally unfamiliar with the fairy tale of Snow White and the Seven Dwarfs.
To understand how the associative thinking principle helps trigger novel ideas, consider connecting the concept of ``apple'' with an unrelated concept, ``paper''. One property of an apple is that it can be peeled. If we apply this property to a paper, we discover that a potential innovation is to make notebook paper with horizontal perforations that allow it be easily torn into a third or a half. This is useful in conserving paper because instead of tearing a whole sheet of paper, only the used part can be torn off and the rest retained for later. While the idea of conserving paper by using smaller portions is not new, as seen in products like paper towel rolls, the innovation here lies in applying this concept to different paper forms, such as notebook paper, inspired by an unrelated natural process—the peeling of an apple. This demonstrates how the simple technique of connecting two concepts that are not typically associated can transform a common idea by cross-applying properties from different domains.

Recently, large language models (LLMs) such as vGPT-4 have shown remarkable results in generating human-like responses in a wide variety of tasks given just a few examples \cite{brown2020language, bubeck2023sparks}. Several works have praised LLMs to produce highly creative outputs \cite{gilhooly2023ai, stevenson2022putting, gomez2023confederacy, chakrabarty2023art, franceschelli2023creativity}. Unlike the rich associations based on different types of connections in human brains, these models implicitly capture associations between concepts based on the frequency of occurrences of commonly related words. In that sense LLMs also form a limited associative network that connects different concepts. What would happen if the models were probed to form associations between words that do not frequently occur together? We hypothesize that \textbf{associative thinking techniques that improve human creativity will also improve LLM creativity.} 

In this paper, we aim to study whether compelling LLMs like vGPT-4 to make connections between unrelated concepts, boosts its creativity. We present three practical domains, Product Design, Storytelling and Marketing to benchmark vGPT-4's performance. For each domain, we devise creativity tasks and score its performance based on originality and usefulness. While we only use vGPT-4 in our experiments, we believe that these results will generalize to other LLM models, given their common underlying associative nature. Overall, we find that prompting vGPT-4 to incorporate associative thinking significantly improves novelty of responses.
\begin{figure}
\begin{center}
\includegraphics[width=\columnwidth]{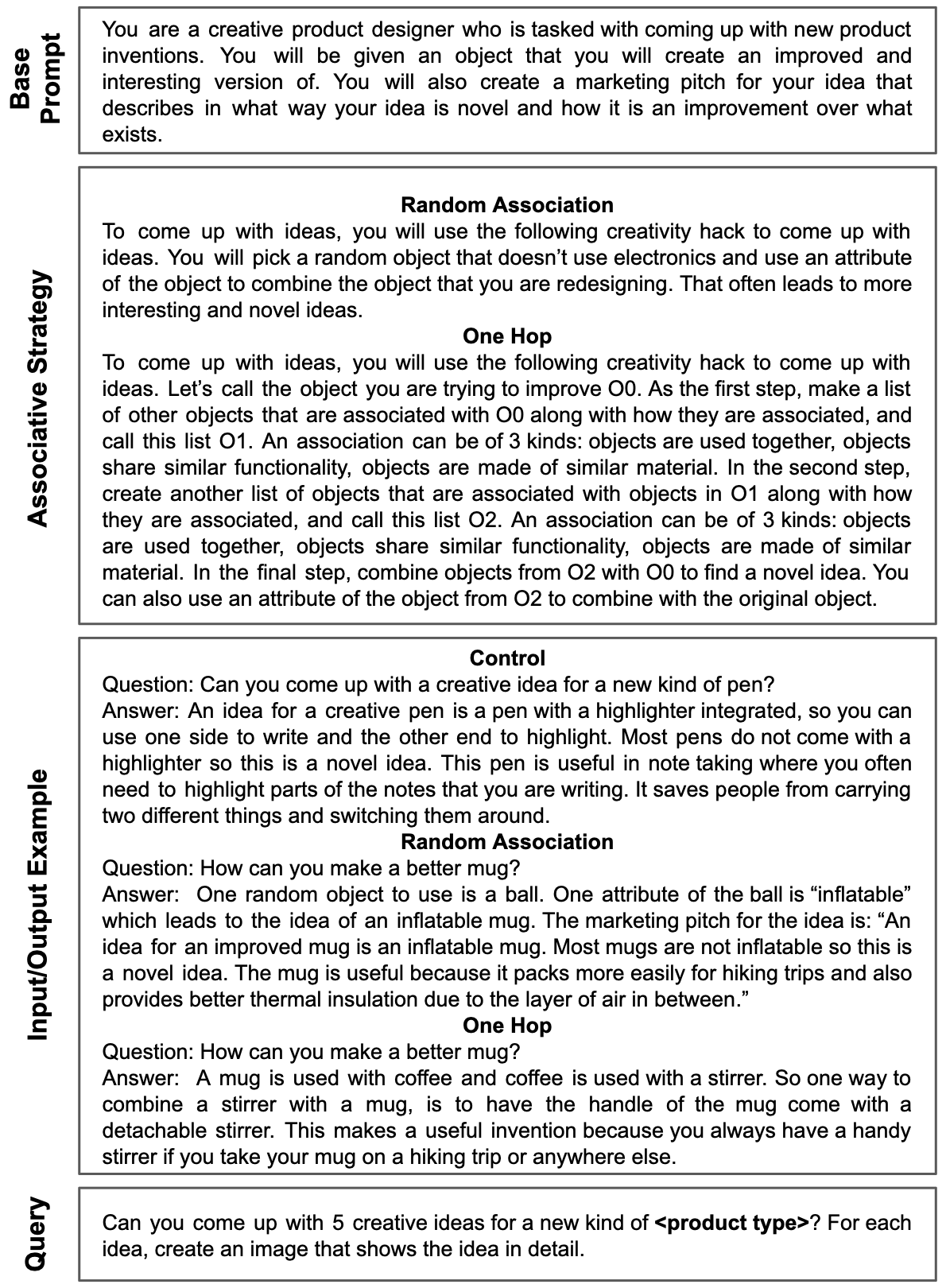}
\end{center}
\vspace*{-5mm}
\caption{The prompt used for applying associative thinking strategy in the Product Design Domain. The prompt begins and ends with the base prompt and query respectively, while we vary the associative thinking strategy based on the condition.} 
\label{productprompt}
\end{figure}

\section{Related Work}
At a high level, creativity can be described as the production of artifacts that are both original and useful, across domains. An original idea that doesn't serve any purpose is merely imaginative, not creative. Evaluations of creativity include these two primary metrics of originality and usefulness \cite{sternberg1999a, Runco2012a, Diedrich2015a} and sometimes include a third factor of style \cite{Besemer1998a} or wholeness \cite{Henriksen2015a}. In this paper, we focus on originality and usefulness as our core metrics. Since human evaluations are influenced by the overall style or design, the third factor gets implicitly factored into the results.    

Creativity has been studied from various perspectives. One framework for understanding its origins is the 4P model \cite{Rhodes1961a}, which identifies four factors: Person (e.g. personality traits), Process, Press (as in environment) and Product (the created artifact). In this paper, we will focus solely on the Product aspect—assessing whether artifacts created by vGPT-4 can be considered creative. This emphasis is practical, as the end product is often what matters most in real-world applications. 

Additionally, creativity can also be analyzed in terms of its magnitude. The 4C model categorizes creativity into several distinct levels: Big-C creativity (eminent levels of creativity like that of Einstein or Mozart), Pro-C (professional level of expertise in a creative domain), little-c (everyday creativity like coming up with a new recipe) to mini-c (an idea that is personally creative even if it exists elsewhere) \cite{Kaufman2009a}. Our focus in this paper lies in the range of little-c and Pro-c levels of creativity, where AI can provide meaningful help to practitioners in different creative domains.

Several works have been proposed to constrain the behavior of LLMs to produce more precise systematic outputs \cite{wei2022chain, nye2021scratchpad, yao2023tree}. These studies focus on convergent thinking, where the goal is to find a single optimal solution. In contrast, creativity entails divergent thinking, where a problem can have multiple solutions. Several recent studies have pitted humans against LLMs on divergent thinking tasks and found that LLMs fare comparably or even better than humans \cite{Stevenson2022a, Cropley2023a, Hubert2023a}. Performance on divergent thinking tests is evaluated by fluency (total number of ideas), flexibility (distinct categories of ideas), originality (based on frequency of other responses) and elaboration. LLMs have access to much more data than humans, giving them a large, unfair advantage in divergent thinking tasks. For example, fluency is trivial for LLMs to produce with their vast store of knowledge. In this paper we adopt a stricter product-focused criteria which directly evaluates creativity based on whether a particular artifact produced by vGPT-4 is creative or not, and whether applying associative thinking can further improve vGPT-4's creative output.

\section{Applying Associative Thinking Strategies to LLMs}
The associative thinking process is defined as \textit{``the forming of associative elements into new combinations which either meet specific requirements or are in some way useful. The more mutually remote the elements of the new combination, the more creative the process or solution''}\cite{Mednick1962a}. 
To enable associative thinking, we prompted vGPT-4 to integrate a random object, or an attribute of the object, in its creative output. Below we describe in more detail each domain, the task given to vGPT-4 and the associative thinking prompts used.

\subsection{Domains}
We present three hand-crafted task domains that have useful applications in the real world, \textbf{Product Design, Storytelling} and \textbf{Marketing}.

The \textbf{Product Design} domain consists 25 everyday objects such as pen, canvas, soap dispenser and pillow. The task here is to come up with creative ideas that improve a given object, as shown in the control prompt for vGPT-4 in Figure \ref{productprompt}. Our associative thinking prompt evaluates two strategies: (1) association with a random object, and (2) association by one-hop method. In one-hop, the goal is to combine the original object with another object by skipping over the first layer of associations. The middle two boxes in Figure \ref{productprompt} show the associative thinking prompts used.

\begin{figure}
\begin{center}
\includegraphics[width=\columnwidth]{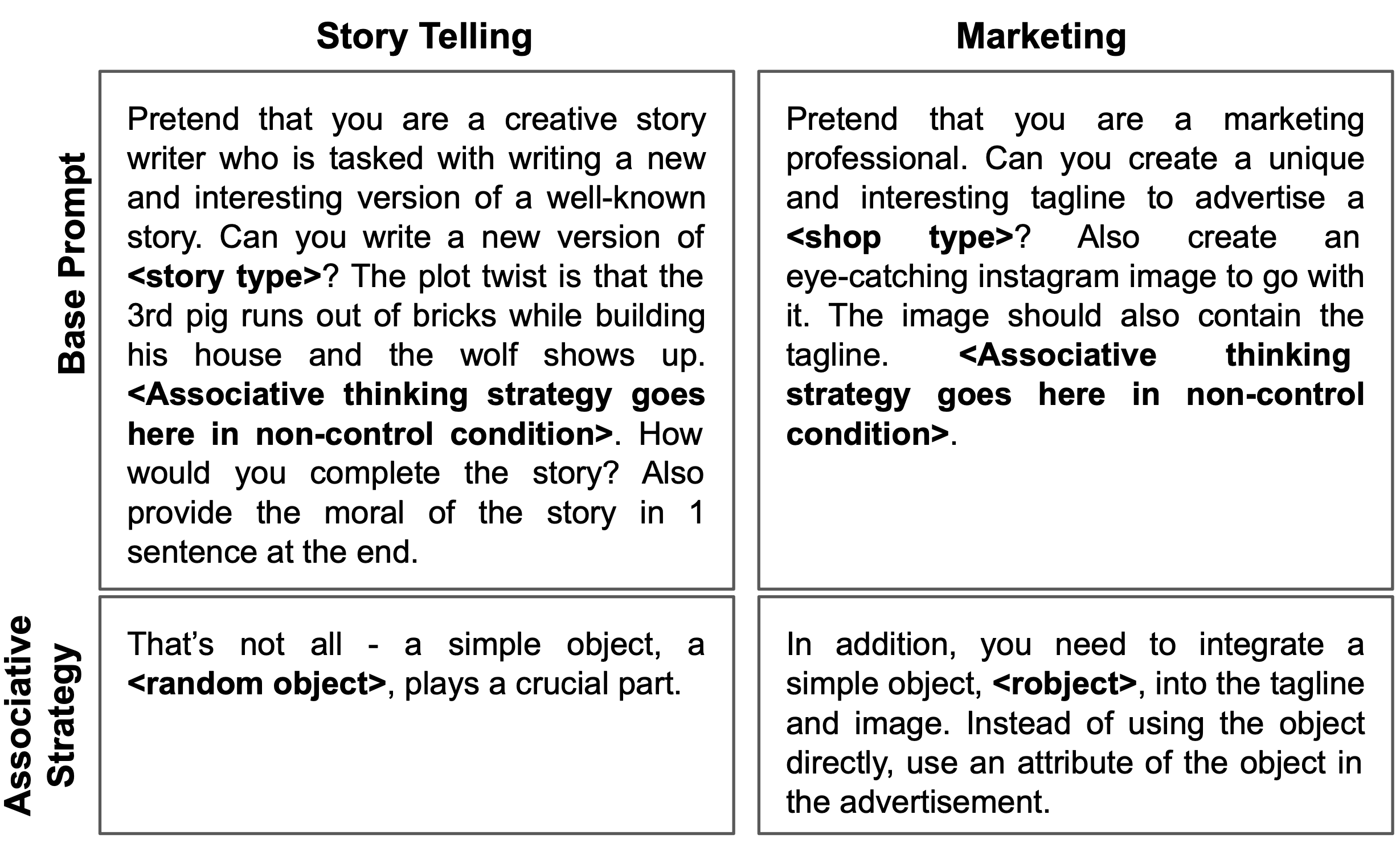}
\end{center}
\vspace*{-5mm}
\caption{Prompts used for the Storytelling and Marketing domains. In the control condition, no associative thinking strategy was introduced. In the Story Telling Domain (left), the story type and plot twist were provided and in the Marketing Domain (right) the shop type was provided. In both cases, we fix a random object and manipulate the associative thinking strategy.} 
\label{story_marketing_prompt}
\end{figure}

In the \textbf{Storytelling} domain, we include 15 popular fairy tales like \textit{Goldilocks and the Three Bears}, \textit{Three Little Pigs}, and \textit{Hansel and Gretel}. In the control condition, we provide a simple twist to the story and ask the LLM to generate a creative version of it. In the associative thinking condition, we additionally ask vGPT-4 to integrate a random object such as a book, a tooth brush or a ladder into the story in a meaningful way. Figure \ref{story_marketing_prompt} (left) presents the prompts we used to test these conditions. By starting with a twist in both versions, we constrain the LLM to stay in the same genre as the original version so that we can draw a fair comparison between both versions. Without this, we found that vGPT-4 tends to generate stories in very different genres like modern or science fiction.

Lastly, in the \textbf{Marketing} domain we introduce 15 small businesses like coffee shop, yoga studio and comic book store. The task here is to create
social media images with a tagline, to advertise these stores. The control prompt (Figure \ref{story_marketing_prompt}, right) says just that. In our associative-thinking prompt, we ask vGPT-4 to additionally incorporate
random objects like a mirror, a bucket and a shovel into the tagline and image (Figure \ref{story_marketing_prompt}, right).  

\subsection{Evaluating Creativity of LLM Output}
To evaluate artifacts created by AI, we use the two core metrics of creativity -- originality and usefulness. Originality measures whether the artifact is novel or not, and usefulness measures whether the artifact solves a meaningful problem or is appropriate in a situation. We adjust these metrics slightly with respect to each domain. We recruited 6 annotators to evaluate v-GPT4 responses. Each scoring task was evaluated by three annotators and the results averaged to get the final scores. With this approach, if 2 out of 3 annotators marked a response as more creative, the final score gets counted as more creative. In the rest of this section, we detail how each domain was evaluated for creativity. 

\textbf{Product Design} 
We quantify originality by assessing whether the idea suggested by vGPT-4 matches with any product already available through sites like Amazon, Etsy, and eBay. If a matching product is available, the originality score is 0, else it is 1.  

To assess usefulness, we evaluate an idea's impact based on whether it addresses a meaningful problem for the target audience. Ideas that focus primarily on aesthetics rather than solving significant issues receive a score of 0. Conversely, ideas with high impact are awarded a score of 1.

\textbf{Storytelling} We evaluate the originality of responses generated by vGPT-4 by how much the story deviates from the original version without disrupting the main story arc. Stories that had major changes to the story arc (different conflict, resolution or ending) got a score of 1 while the others got a score of 0 (low originality). 

We measure usefulness of a story by its ability to engage the reader. To assess usefulness, we set the outputs from the control and associative thinking conditions as two versions of the story, A and B. To eliminate bias towards version B, we randomly shuffle the order in which the control story appears as A or B. We then present the two versions to our annotators and ask them to pick the one they find more interesting.  

\textbf{Marketing}
To evaluate creativity for marketing advertisements, we present our human annotators with both the control and associative thinking versions. Similar to the Storytelling domain, we randomly shuffle the order in which the control appears as version A or B. To assess originality, we ask the annotators to choose the version which one seems less typical for its category, and to assess usefulness we ask them to choose the version they find more persuasive as an advertisement. 

\subsection{Experiments}
\begin{figure}
\begin{center}
\includegraphics[width=\columnwidth]{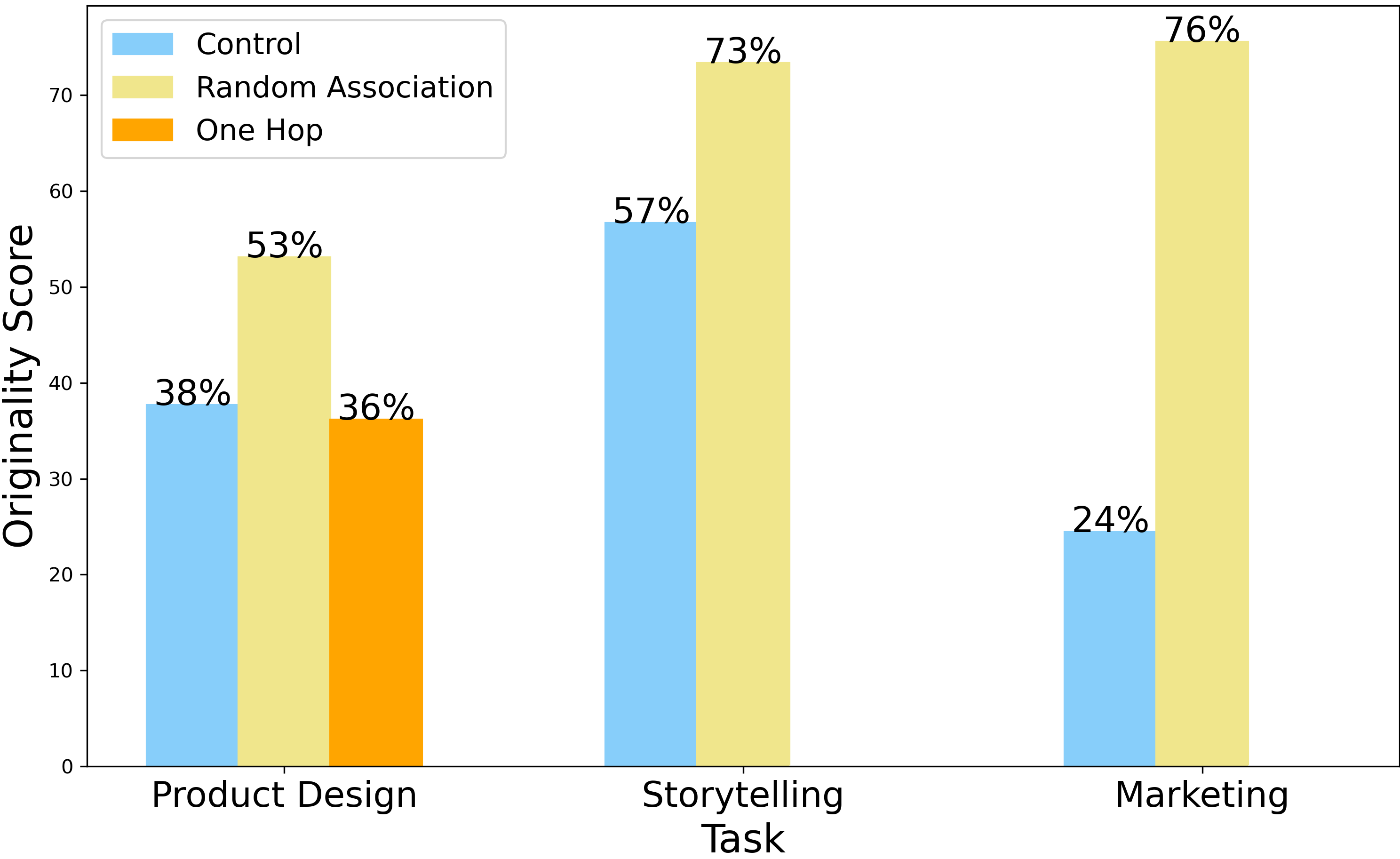}
\end{center}
\vspace*{-5mm}
\caption{Originality scores for Product, Storytelling and Marketing Domains. Overall, the Random Association Associative thinking strategy boosts originality across all domains.
} 
\label{orignality_score}
\end{figure}

\begin{figure}
\begin{center}
\includegraphics[width=\columnwidth]{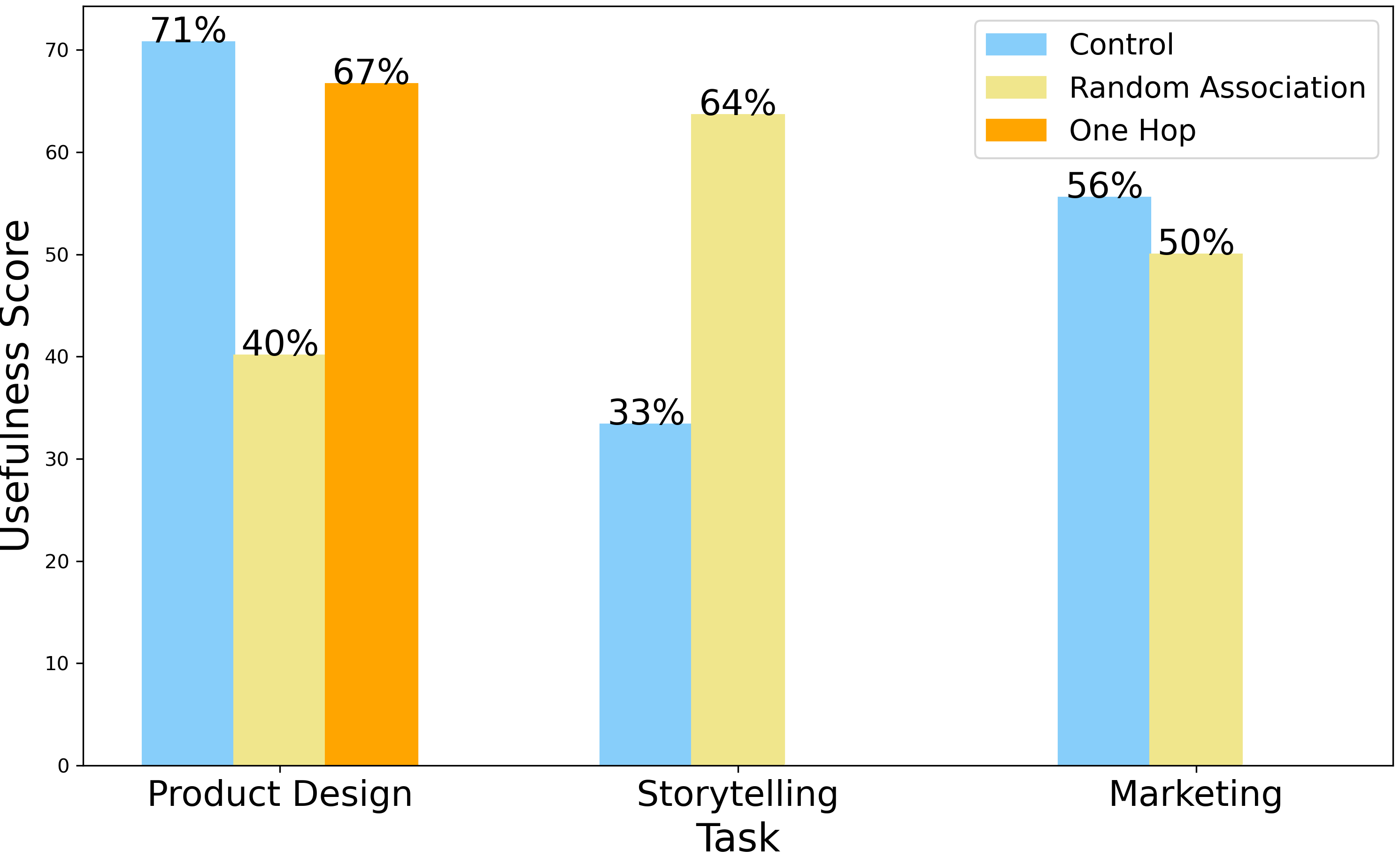}
\end{center}
\vspace*{-5mm}
\caption{Usefulness scores for Product, Storytelling and Marketing Domains. Associative thinking strategy results in more useful outcomes in Storytelling and Marketing domains but appears to be less useful for the Product Design domain. 
} 
\label{usefulness_score}
\end{figure}

To test our hypothesis, we gave vGPT-4 creative challenges in designing improved versions of products, writing creative stories and designing marketing advertisements. We used the prompts described in figures \ref{productprompt} and \ref{story_marketing_prompt} with the default temperature setting of 1. Across all the domains, we test two conditions, a control condition and an associative thinking condition with explicit instructions and input/output examples explaining how to integrate random objects into the response. For the Product Design Domain, we test the one-hop associative thinking method as an additional variant. Finally, we obtain human judgement scores for originality and usefulness by assigning three unique annotators per task per domain. Figures \ref{orignality_score} and \ref{usefulness_score} show the originality and usefulness scores of responses generated by vGPT-4 across all the three domains. 

Integration of unrelated concepts leads to originality, but it comes at a cost as the novel combination might appear incongruous. We applied two strategies to reduce incongruity where appropriate: (1) using an attribute of the random object instead of the object directly, and (2) reducing the randomness of the stimulus using the one-hop method. We applied both (1) and (2) to the Product Design domain, and (1) to the Marketing domain.

\section{Results}
We found that associative thinking by random association improved originality of responses across all three domains. 

\subsection{Product Design}
Associative thinking by \textit{random association} improved the originality score by 40\% compared to the control. However, random association ideas were rated less useful than the control scenario, which reflects the challenge of integrating unrelated elements in a meaningful way. Associative thinking using the \textit{one-hop method} was similar to the control version for both originality and usefulness.  
\begin{figure}
\begin{center}
\includegraphics[width=\columnwidth]{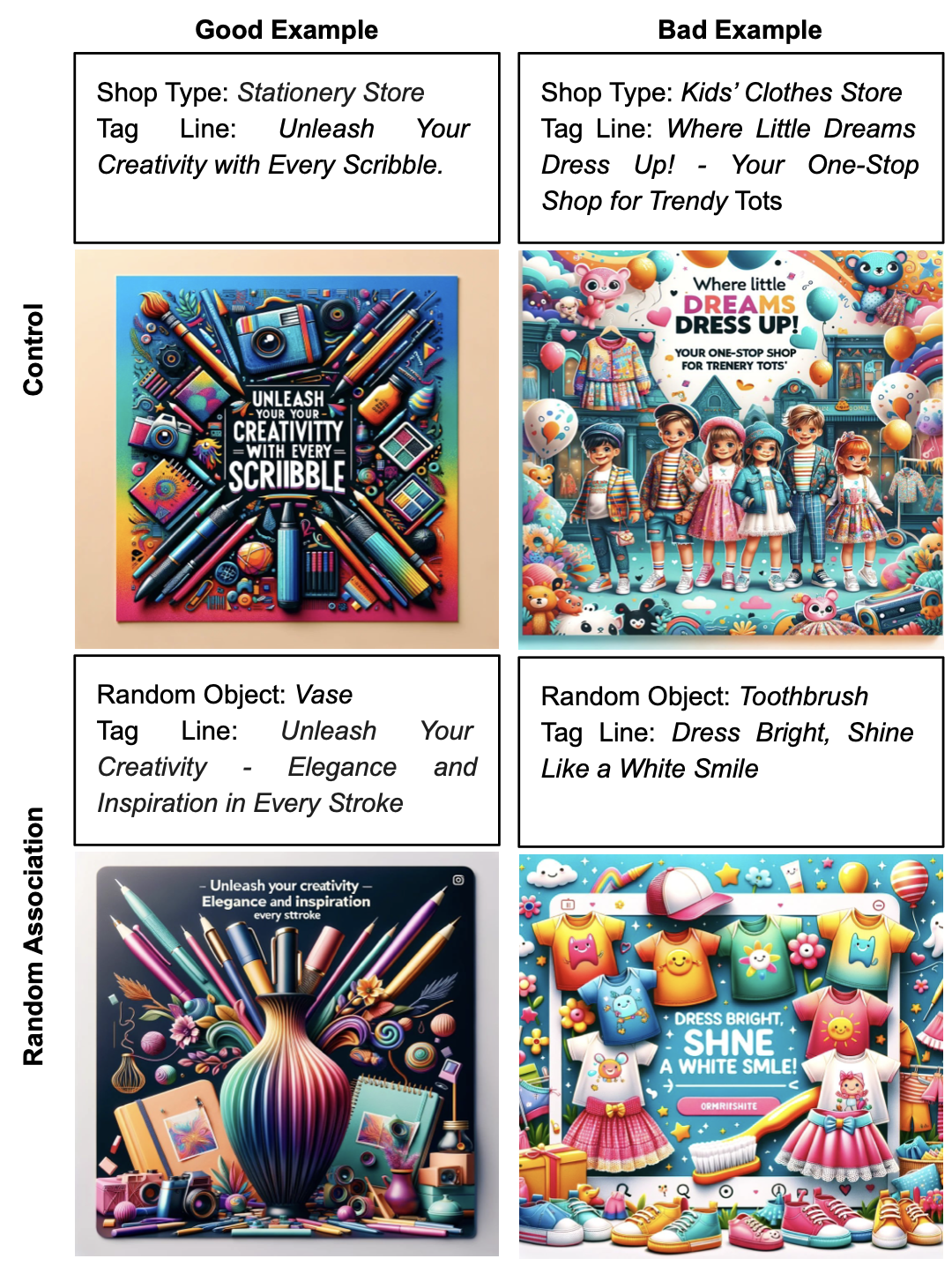}
\end{center}
\vspace*{-5mm}
\caption{Responses from vGPT-4 for the Marketing Domain. For many shop types, we found the results to be quite persuasive and useful, like the left one. However, sometimes the LLM would output images where the random object did not integrate in a useful way, like the example on the right.} 
\label{marketing}
\end{figure}

\begin{figure}
\begin{center}
\includegraphics[width=\columnwidth]{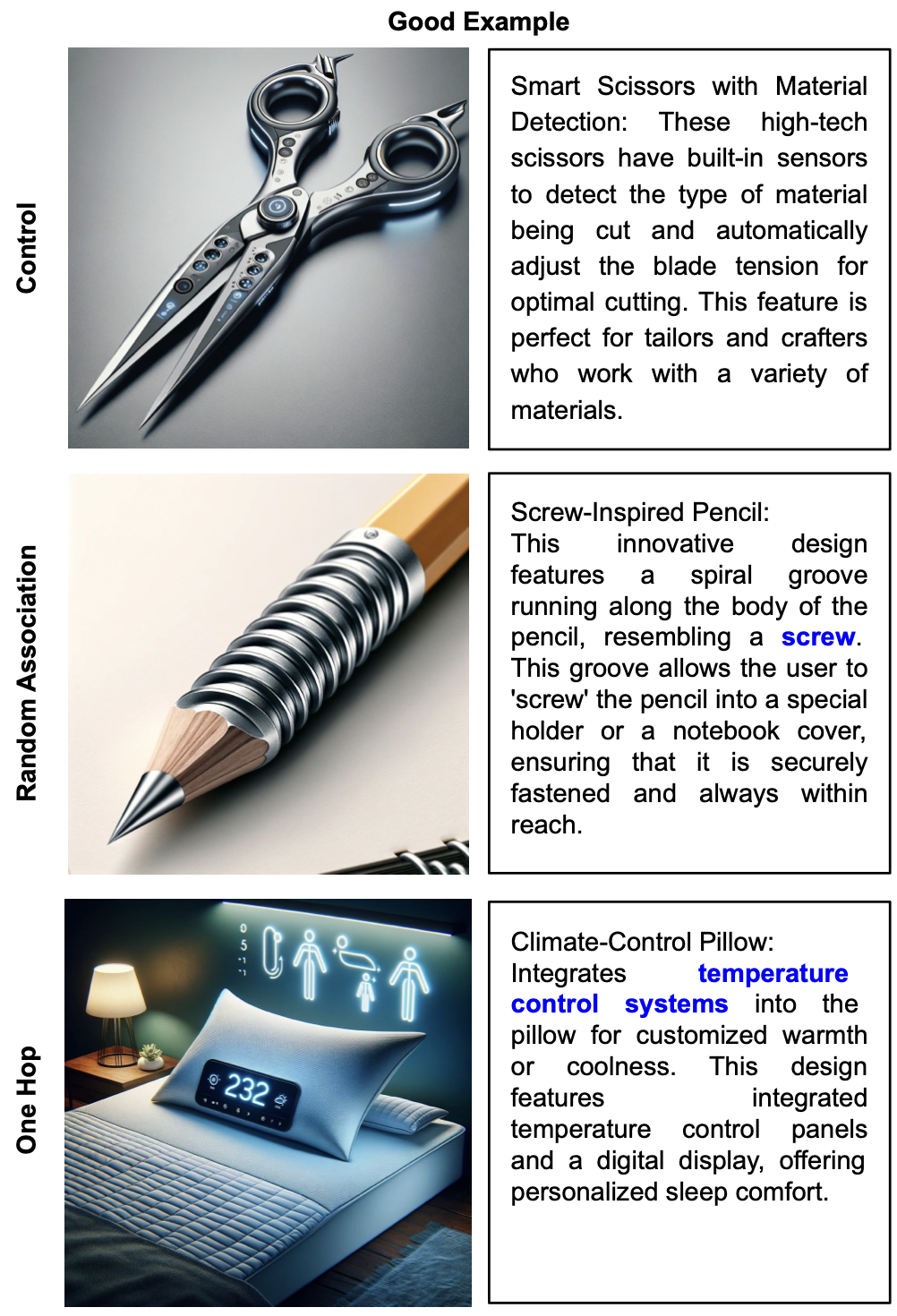}
\end{center}
\vspace*{-5mm}
\caption{Instances where vGPT-4 generated novel and useful ideas in the Product Domain. These product ideas were scored only for the textual outputs, but we include the images here for reference. We observe that some images make the idea look unworkable, for example, the screw-like groove may prohibit sharpening the pencil. However, a simple tweak like treating the screw-like groove as a pencil sleeve could easily solve the issue. } 
\label{productgood}
\end{figure}

One factor that partly explains the high usefulness score of the control and one-hop versions is that they had a higher percentage of ideas incorporating some form of electronics (43\% and 41\% respectively) compared to random association (23\%). Electronic components add functionality and scored higher on our simpler criteria of usefulness. For example, the Smart Scissors with Material detection (Figure \ref{productgood}) scores high on impact but may not be practical as a business idea. 

\begin{figure}
\begin{center}
\includegraphics[width=\columnwidth]{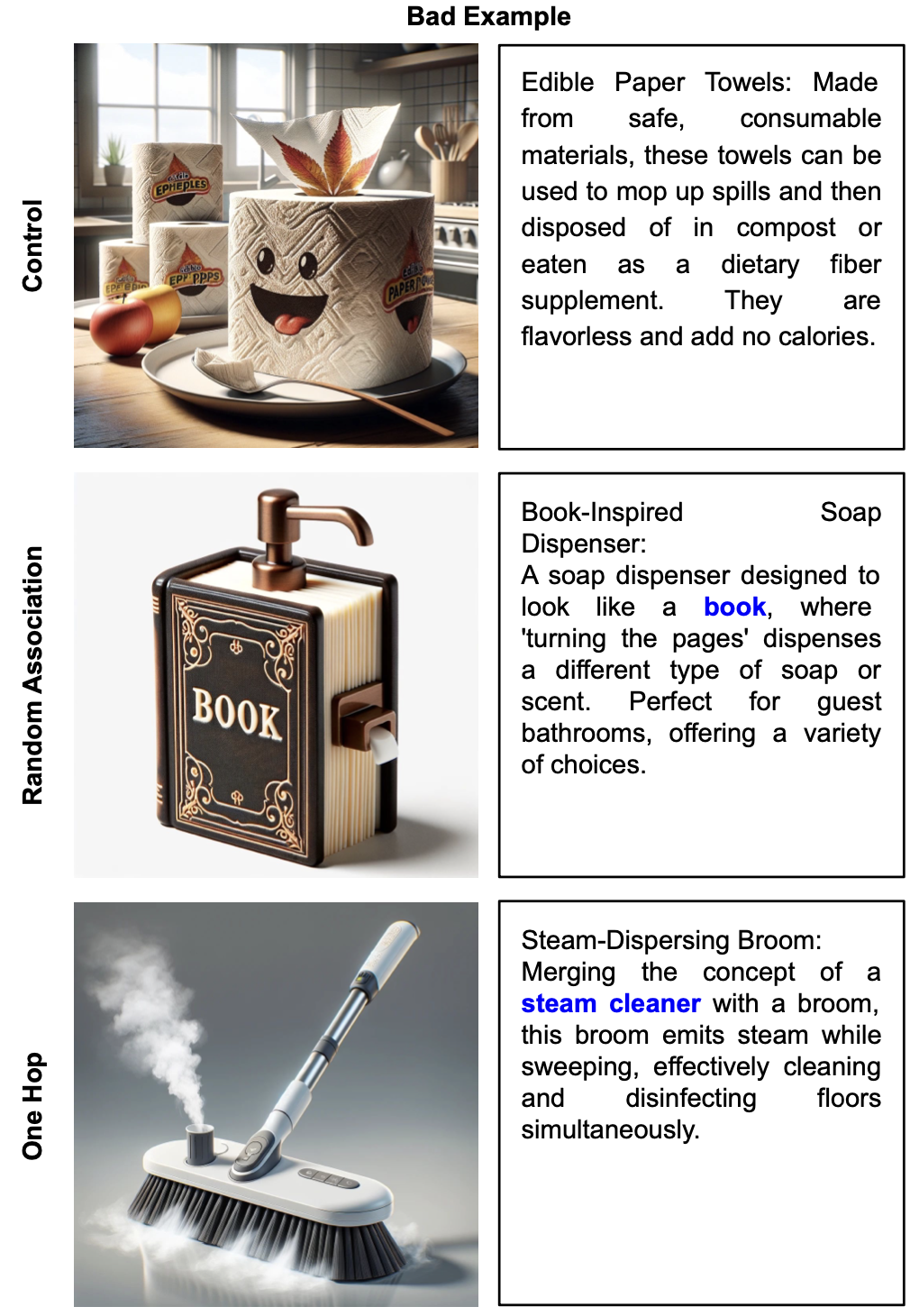}
\end{center}
\vspace*{-5mm}
\caption{Instances for the Product Domain where the generated idea not only fails to solve a meaningful problem but is also undesirable. For example, paper towel used to clean up spills is undesirable as an edible item. Moreover, brooms, which are originally meant to clean dry dirt could cause the dirt to spread further if incorporated with steam.
} 
\label{productbad}
\end{figure}

The idea behind the one-hop strategy is to reduce incongruity by connecting ideas that are not directly associated but also not completely unrelated. It achieves this by skipping over the first level of associations from the original object.  However, we found that vGPT-4 picks final associations that are closely related to the original object, thereby reducing this scenario to the control version. For example, it goes from \textit{comb} $\rightarrow$ \textit{hair gel} $\rightarrow$ \textit{styling mousse}, or from \textit{scissors} $\rightarrow$ \textit{knife} $\rightarrow$ \textit{cutting board}. Since there is reasonably strong association between \textit{comb} $\rightarrow$ \textit{styling mousse} and \textit{scissors} $\rightarrow$ \textit{cutting board}, the one-hop approach does not lead to any meaningful associative thinking. 
Figure \ref{productgood} and \ref{productbad} are few selected product ideas generated by vGPT-4. 

\subsection{Storytelling}
\begin{figure*}
\begin{center}
\includegraphics[width=18cm]{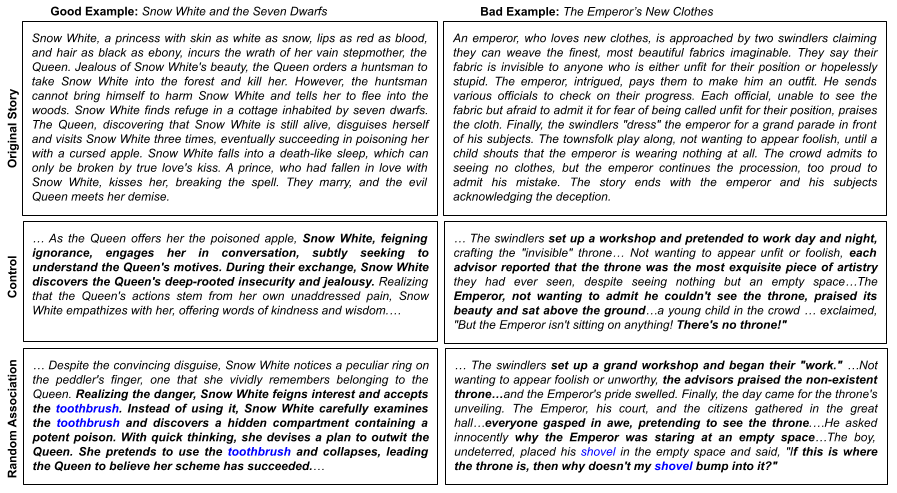}
\end{center}
\vspace*{-5mm}
\caption{Stories generated by vGPT-4. While many stories, like the left one, changed the main plot or ending, we also found that the LLM generated stories that did not deviate much from the original story.} 
\label{story}
\end{figure*}

Associative thinking improved the originality of stories by 28\% over the control scenarios. In addition, associative thinking stories were twice as likely to be rated more interesting, possibly due to the random object causing an increase in conflict part of the story arc. When we encounter a conflict in a story, we release small amounts of the stress hormone, cortisol, which increases our focus and keeps us engaged in the story \cite{Srivastava2023a}. The example in Figure \ref{story} shows the difference in the two versions of Snow White and the Seven Dwarfs, where the initial twist was that Snow White recognizes the evil Queen when she comes to poison her. In the version that uses a toothbrush (shown in blue) as a random element, the conflict part (shown in bold) gets extended making the story more interesting.  

\subsection{Marketing}
We found a significant improvement in the originality score of advertisements with the inclusion of the random object. As an example, for the Sports Memorabilia Store advertisement, that had to integrate a broom, vGPT-4 came up with the slogan, ``Sweep Up History's Greatest Moments!'' a tagline it would likely not have arrived at without the explicit suggestion of broom. With a clever integration, vGPT-4 also flipped the typical negative connotation of broom into a positive one.  

Overall, we found that the advertisements generated under the associative thinking condition received a lower usefulness score than the control condition. In cases where the usefulness was high, the inclusion of the random object focused vGPT-4 towards a distinct theme, leading to a simpler message that felt intentional. For example, in Figure \ref{marketing} (bottom-left), the addition of the vase creates a distinct artistic vibe that feels cleaner compared to the control version (top-left) which feels crowded. 
In some situations, inclusion of the random element appeared incongruous and rated lower on persuasiveness. For example, when combining a toothbrush with a Kid's Clothes store, the the juxtaposition of toothbrush with cute clothes appears jarring (bottom-right in Figure \ref{marketing}).

\section{Limitations}
We outline two broad limitations encountered during our experiments that can guide more research in this area. 

\textbf{Context Dependency:} In one-third of the storytelling prompts, vGPT-4 gave magical properties to the random object (e.g. in the Aladdin story, the genie transforms the curtain into a magical item that brings peace when draped over someone). While the use of magic is appropriate for fairy tales, it also leads to less surprising and less satisfying resolutions (magic can fix any problem!). Even when prompted to list attributes of the object as used in everyday life, vGPT-4 still gave the object magical properties (e.g. ``mystical patterns’’ and ``memory keeper’’ for the curtain). In contrast, this did not happen in the Product Design or the Marketing scenarios, implying a high context dependency on how the object is used. Many creative ideas come from applying concepts from one domain to another, but it requires us to keep the two contexts independent and switch between them seamlessly. While context awareness is extremely important in generating relevant responses, not being able to switch contexts limits creativity.

\textbf{Associative Rut:} In the product domain scenario, vGPT-4’s responses show an over representation of nature inspired random objects, such as tree and leaf. This is similar to the phenomena of ``associative rut’’ which is a common occurrence during creative problem solving \cite{Santanen2004a}. Associative rut occurs when a particular concept leads to the activation of dominant links that bring forth already known ideas to the consciousness. vGPT-4’s preference for trees and leaves to inspire product improvements reflects the popular ethos of eco-friendly designs in the product domain. This is different from context dependency above, as even within the same domain vGPT-4 uses only a small subset of concepts repeatedly.

\section{Conclusion}
Our aim was to asses whether associative thinking strategies can help LLMs like v-GPT4 boost their creativity. We presented three domains, Product Design, Story Telling and Marketing. For each domain, we crafted prompts that guided v-GPT4 to apply associative thinking. We tested these prompts under two conditions, the first condition served as a control, without any associative thinking and the second condition explicitly incorporated associative thinking strategy. We assessed the creativity of v-GPT4s responses using human judgements for their originality and usefulness. Overall, we found that the \textbf{associative thinking strategy resulted in more original responses across all domains}. Usefulness of the responses varied across domains. In Story Telling, associative thinking helped generate more useful stories but not in the Product Design and Marketing domains. Yet, a usefulness score over 40\%, makes roughly one in two ideas proposed by vGPT-4 worth considering. 
In general, evaluating usefulness of an idea is tricky because many original ideas may initially sound redundant but can lead to better ideas with some rework. As an example, one Product Design idea generated by random association that all annotators uniformly rated low was ``\textit{Origami-Inspired Foldable Newspaper}'' that could be transformed into a portable reading stand. While this idea itself may not be as practical, the concept of origami along with soon-to-be-recycled paper could potentially lead to other ideas. For example, marketing flyers with faint origami lines would encourage people to read the flyer while doing a fun activity with their kids. 
Our results show that associative thinking can be a useful tool for making vGPT-4 more creative. Additionally, vGPT-4 can be a useful tool especially in the initial phases of brainstorming, even in challenging domains like Product Design. 

\section{Acknowledgements}
We thank our volunteers who supported the scoring of our experimental results.
\bibliographystyle{apacite}
\setlength{\bibleftmargin}{.125in}
\setlength{\bibindent}{-\bibleftmargin}
\bibliography{references}
\end{document}